\documentclass[a4paper, 10pt, conference]{IEEEtran}      %Use this line for a4
\IEEEoverridecommandlockouts                              % This command is only
\usepackage{multirow}
\usepackage{multicol}
\usepackage{supertabular}
\usepackage{longtable}
\usepackage{lscape}
\usepackage[utf8]{inputenc}
\usepackage[T1]{fontenc}
\usepackage{graphicx}
\usepackage{wrapfig}
\usepackage{float}
\graphicspath{ {./images/} }
\usepackage{booktabs}
\usepackage{mathptmx} % assumes new font selection scheme installed
\usepackage{mathptmx} % assumes new font selection scheme installed
\usepackage{amsmath} % assumes amsmath package installed
\usepackage{amssymb}  % assumes amsmath package installed

\title{\LARGE \bf
Machine Learning Based Network Coverage Guidance System
}

\author{Srikanth Chandar, Muvazima Mansoor, Mohina Ahmadi, Hrishikesh Badve, Deepesh Sahoo, Bharath Katragadda$^{1}$
\thanks{$^{1}$Third-year B-Tech ECE students, PES University, Outer Ring Rd,
Banashankari 3rd Stage, Banashankari, Bengaluru, Karnataka 560085

        {\tt\small Srikanth.chandar@gmail.com,\newline    muvazima99@gmail.com, mohina1729@gmail.com,\newline hrishi.badve11@gmail.com,\newline    deepeshsahoo99@gmail.com, \newline ashubharath99@gmail.com,  }}%

}

\begin{document}
\maketitle
\thispagestyle{empty}
\pagestyle{empty}

%%%%%%%%%%%%%%%%%%%%%%%%%%%%%%%%%%%%%%%%%%%%%%%%%%%%%%%%%%%%%%%%%%%%%%%%%%%%%%%%
\begin{abstract}
With the advent of 4G, there has been a huge consumption of data and the availability of mobile networks has become paramount. Also, with the burst of network traffic based on user consumption, data availability and network anomalies have increased substantially. In this paper, we introduce a novel approach, to identify the regions that have poor network connectivity thereby providing feedback to both the service providers to improve the coverage as well as to the customers to choose the network judiciously. In addition to this, the solution enables customers to navigate to a better mobile network coverage area with stronger signal strength location using Machine Learning Clustering Algorithms, whilst deploying it as a Mobile Application. It also provides a dynamic visual representation of varying network strength and range across nearby geographical areas.

\end{abstract}

%%%%%%%%%%%%%%%%%%%%%%%%%%%%%%%%%%%%%%%%%%%%%%%%%%%%%%%%%%%%%%%%%%%%%%%%%%%%%%%%
\section{INTRODUCTION}
The growing demand for mobile network connectivity associated with
increased smartphone ownership, greater mobile usage indoors and higher
data rates are driving the evolution of mobile networks.
Due to the annual increase in the number of cellular subscribers and the increase in competition with other network operators, there is a growing interest by the network operators to maximize the deployment of the network infrastructure to achieve maximum coverage. The definition of network infrastructure is not only limited to electronic components of the network but also the passive elements such as physical sites and towers that are required to operate the network. While the cellular subscribers do not directly perceive the composition or the configuration of the infrastructure, the throughput and latency of the mobile network infrastructure determine the user experience and therefore the network infrastructure and its deployment have been one of the key challenges faced by Mobile Network Operators (MNOs). Since network coverage is of utmost importance to the MNOs, one would expect the users to have a seamless experience with strong uniform network connectivity. However, this is not the case. The network coverage distribution in tall structures /large buildings is inconsistent and significant variations in signal strengths exist. With the increase in dependency on network connectivity and the need for high network speeds, the lack of consistent and strong signal strength is a growing concern. 

In this paper, a unique approach to resolve the problems faced by the MNOs (identification of areas with weak signal strengths), as well as the problems faced by end-users (weak signal strengths), is presented. With the help of a mobile application, network strength densities across a region are identified and 360-degree feedback on network conditions to both MNOs and end customers is provided. The solution presented in this paper enables the end-user to navigate to a location with a stronger mobile network and also provides a dynamic visual representation of varying network strength and range across nearby geographical areas. The dynamic visual representation of varying network strength is provided for the different kinds of MNOs that operate in a region. This enables the user to judiciously switch to an MNO with more uniform signal connectivity and also provides the MNOs a comparative view of the signal strengths of their competition.

\section{Related Works}
 Parallels can be drawn between this paper and the work done in Opensignal\cite{ref11}, in terms of deploying network strength-based heat-maps. The means of deploying heat-maps in Opensignal is one through a meticulous process of data collection and averaging. It deals with collecting billions of individual measurements every day, from over 100 million devices worldwide, per day. In addition to reaching such extensive limits of data collection, there exists a dependency on partner applications to collect said data, which aids the data collecting venture. An averaging metric is then applied onto this vast database, to produce heat-maps pertaining to a geographical location, and of a particular MNO. While this may be accurate owing to the sheer quantity of data collected, using a clustering model to attribute new data into pre-defined clusters and periodically updating the cluster metric itself, will definitely reduce complexity and be cheaper. Using such algorithms to fill in the blanks of the existing vast geography may be more efficient than trying to get every location's detail, which essentially amounts to a brute force method. Another outlook is to use such algorithms on existing vast databases- such as the ones in Opensignals- which may be further added to accuracy while reducing complexity and cost.
\section{PROBLEM STATEMENT}

\subsection{Poor Network Connectivity}

The mobile network user faces the issue of poor internet connectivity multiple times in a day, in a particular area, due to various reasons – most of which are due to external factors such as physical obstruction, multiple reflections due to water bodies or tank chambers, weather at the given instant, etc. But more often than not, this network strength magnitude is not constant over an area; it is not even constant across the same building. There are points in the building where there is an evident increase in network strength. If there were a tangible way to deduce such geographical locations and direct the user towards the same, much of today’s day to day network-related issues could be solved to a good extent, with ease. For instance, by directing the user to a window, from inside a lift.

\subsection{Telecommunication companies}
Acquiring sites for the deployment of network infrastructure has become very difficult due to network densification to address demands in indoor environments. Due to closely spaced buildings, there is very little space for indoor base stations to be installed. Furthermore, multiple mobile operators have to compete for the same few sites.
The comparative data of signal strengths of all Mobile Network Operators in a particular area could help the telecommunication companies to judiciously deploy network infrastructure or participate in infrastructure sharing. 

\section{High Level Solution}
\begin{figure}[H]
      \centering
      \includegraphics[scale=0.17]{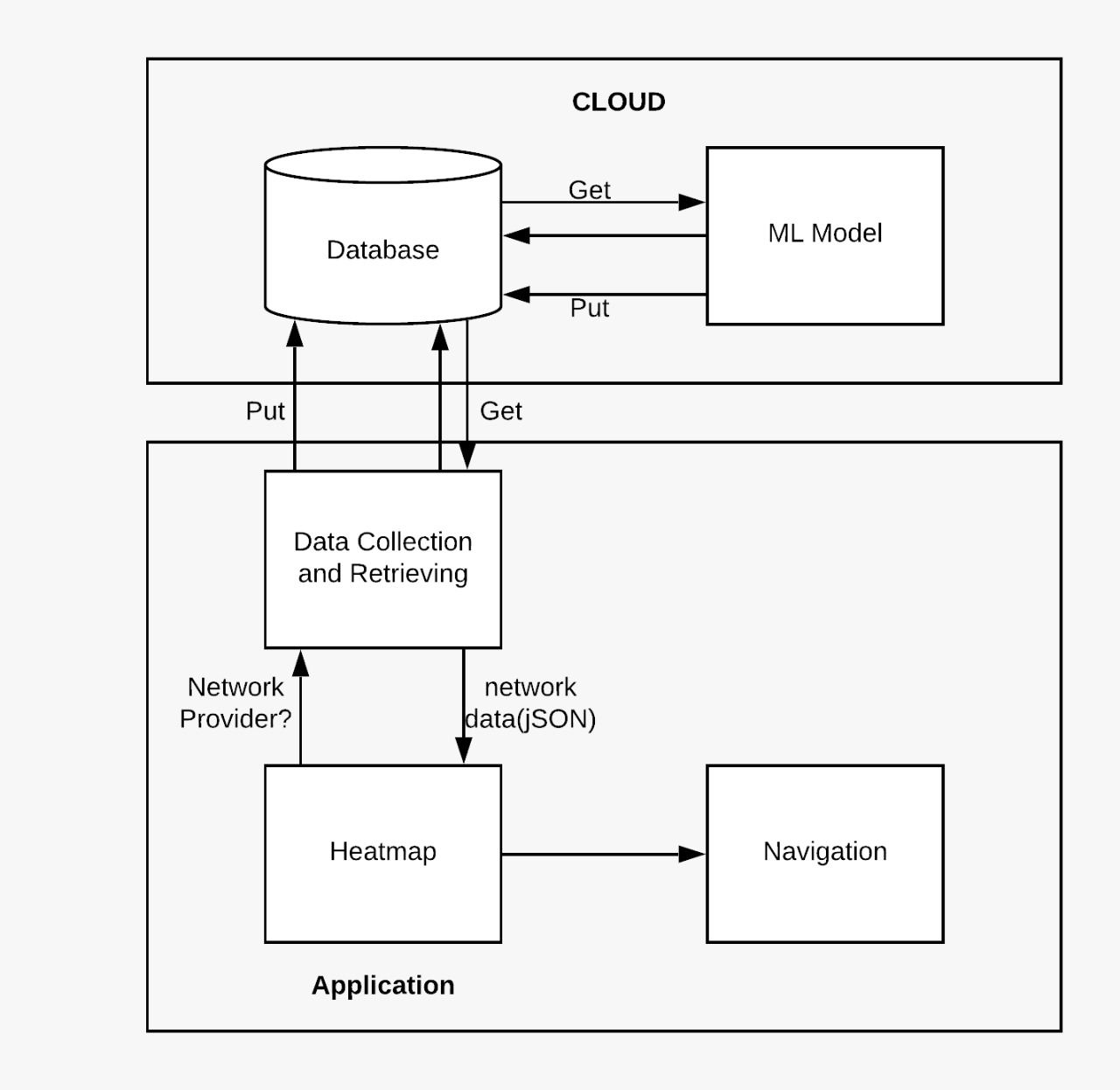}
      \caption{ High-Level Heat map Architecture Diagram}
      \label{figurelabel1}
   \end{figure}
This white paper proposes a solution to the above-mentioned problems, by using an interactive mobile application front that would cater to the user and the user’s geographical area’s network problems; while simultaneously building the database in the back end, which could provide greater insight into larger signal network problems as a whole. 
 The solution is a mix of data engineering, vector quantization algorithms, heat maps, geographical area identifications to partition User Endpoints (UEs) based on - geographical location, Long-Term Evolution Received Signal Strength Indicator (LTE RSSI) and Mobile Network Operator (MNO). The vector quantization approach generates clusters and each cluster is associated with an integer tag, thereby normalizing and denoting the respective network strength of that cluster. The clustering model accuracy increases as more UEs start consuming the RSSI mapping service and generate data to build the clustering model.  Currently, a window of 10 seconds is used to generate the data. Once the model is built and if the UEs fall under a lower network connectivity area, they are directed to the nearest location, belonging to a region with a stronger network, computed based on optimization algorithms using RSSI of the UEs. The UE is then directed to this new stronger network area, using traditional navigation modules. From a user’s perspective, this framework could serve as an advantageous alternative to current day extant methods which predominantly use a trial and error approach, to compare network signal strengths. On the other hand, the data collected could prove useful to the Network Service providers, as it provides an insight into the continuous gradient of signal strength, across an area. This could help in better frequency planning and cell site deployment upgrades. It can also provide strategic insight into the exact location where a signal tower can be installed, in such a way that it is beneficial to the maximum number of weak network clusters, thereby optimizing cost.

\section{Detailed Solution}
The proposed framework can be divided into 3 phases-

\begin{itemize}

\item Data Collection and Storage
\item Cloud Clustering Model
\item Heat map, Nearest Strong Network Area Navigation
\end{itemize}
\subsection{Data Collection and Storage phase}
The front end mobile application updates network and geographic-specific details from the user, every 10 seconds. This process runs as a Background Service, while simultaneously storing this data into Google-Firebase\cite{ref1}. The Firebase Realtime Database is a NoSQL cloud-based database that syncs data across all clients in real-time and provides offline functionality. Data is stored in the Real-time database as JavaScript Object Notation (JSON), and all connected clients share one instance, automatically receiving updates with the newest data\cite{ref1}.
%A brief description of the type of data collected is discussed below.
%\begin{figure}[H]
%      \centering
%      \includegraphics[scale=0.35]{images/g1.jpeg}
%      \caption{The data obtained after using getAllCellInfo() method }
%      \label{figurelabel2}
%   \end{figure}
\subsubsection{Geographic Location}
The user’s precise geographical location is determined by collecting the latitude and the longitude of the device. This data is collected using the FusedLocationProviderApi\cite{ref2}. Fused Location Provider gives accurate locations, and optimizes the battery usage. It combines signals from GPS, Wi-Fi, cell networks, as well as an accelerometer, gyroscope, magnetometer, and other sensors to provide accurate results. It can demarcate various spots within a building or a household.

\subsubsection{Network Strength}
The effective network strength of the device is a metric of the LTE RSSI data. The carrier RSSI measures the average total received power in the measurement bandwidth over N resource blocks, in decibels. The total received power of the carrier RSSI includes the power from co-channel serving and non-serving cells, adjacent channel interference, thermal noise, etc. It is totally measured over 12-sub carriers including Received Signal (RS) from Serving Cell and Traffic in the Serving Cell.
The module used to collect the RSSI data is getAllCellInfo() method under the class TelephonyManager\cite{ref3}. It requests all available cell information from all radios on the device including the camped/registered, serving, and neighboring cells. The response can include one or more CellInfoGsm, CellInfoCdma, CellInfoTdscdma, CellInfoLte, and CellInfoWcdma objects, in any combination. Refer Table \ref{tab:my-table}.

\bottomcaption{Description of getAllCellInfo() fields}
\begin{supertabular}
{|l|l|l|}

%\caption{Description of getAllCellInfo() fields}
\label{tab:my-table}\\
\hline
OBJECT &
  FIELD &
  DESCRIPTION \\ \hline
%\endhead
%
\multirow{3}{*}{\begin{tabular}[c]{@{}l@{}}CellInfoLte-\\ Immutable cell\\  information \\ from a point\\  in time.\end{tabular}} &
  mRegistered &
  \begin{tabular}[c]{@{}l@{}}Indicates whether \\ the cell is being \\ used or would be \\ used for signalling\\ communication if \\ necessary.\end{tabular} \\ \cline{2-3} 
 &
  mTimeStamp &
  \begin{tabular}[c]{@{}l@{}}To determine the\\  recency of \\ CellInfo data.\end{tabular} \\ \cline{2-3} 
 &
  \begin{tabular}[c]{@{}l@{}}mCellConne\\ -ction Status\end{tabular} &
  \begin{tabular}[c]{@{}l@{}}Determines the \\ connection status\end{tabular} \\ \hline
\multirow{9}{*}{\begin{tabular}[c]{@{}l@{}}CellIdentityLte- \\ to represent a\\  unique LTE cell\end{tabular}} &
  MCI &
  28 bit cell identity \\ \cline{2-3} 
 &
  MPCI &
  Physical cell ID \\ \cline{2-3} 
 &
  MTAC &
  \begin{tabular}[c]{@{}l@{}}16 bit tracking \\ area code\end{tabular} \\ \cline{2-3} 
 &
  MRFNC &
  \begin{tabular}[c]{@{}l@{}}18 bit absolute RF \\ channel number\end{tabular} \\ \cline{2-3} 
 &
  mBandwidth &
  \begin{tabular}[c]{@{}l@{}}Cell Bandwidth\\  in KHz\end{tabular} \\ \cline{2-3} 
 &
  mMCC &
  \begin{tabular}[c]{@{}l@{}}Mobile country\\  code\end{tabular} \\ \cline{2-3} 
 &
  mMNC &
  \begin{tabular}[c]{@{}l@{}}Mobile network \\ code\end{tabular} \\ \cline{2-3} 
 &
  mAlphaLong &
  \begin{tabular}[c]{@{}l@{}}Network carrier \\ name\end{tabular} \\ \cline{2-3} 
 &
  mAlphaShort &
  \begin{tabular}[c]{@{}l@{}}Network carrier \\ name\end{tabular} \\ \hline
\multirow{8}{*}{\begin{tabular}[c]{@{}l@{}}CellSignalStrength\\ Lte-LTE signal \\ strength related \\ information.\end{tabular}} &
  RSSI &
  \begin{tabular}[c]{@{}l@{}}Received Signal\\ Strength Indication\end{tabular} \\ \cline{2-3} 
 &
  RSRP &
  \begin{tabular}[c]{@{}l@{}}Reference SIgnal\\  Received Power\end{tabular} \\ \cline{2-3} 
 &
  RSRQ &
  \begin{tabular}[c]{@{}l@{}}Reference SIgnal\\  Received Quality\end{tabular} \\ \cline{2-3} 
 &
  RSSNR &
  \begin{tabular}[c]{@{}l@{}}Reference Signal\\ Signal to Noise \\ Ratio\end{tabular} \\ \cline{2-3} 
 &
  CQI &
  \begin{tabular}[c]{@{}l@{}}Channel Quality\\  Indicator\end{tabular} \\ \cline{2-3} 
 &
  TA &
  Timing Advance \\ \cline{2-3} 
 &
  LEVEL &
  \begin{tabular}[c]{@{}l@{}}Level of signal\\  strength\end{tabular} \\ \cline{2-3} 
 &
  OPLEVEL &
  Output Level \\ \hline
\end{supertabular}

\subsubsection{Network Service Provider}
The Network Service Provider- more formally known as Mobile Network Operator- is a provider of wireless communications services that owns or controls all the elements necessary to provide services to an end-user including radio spectrum allocation, wireless network infrastructure, backhaul infrastructure, billing, customer care, provisioning computer systems, and marketing and repair organizations. 
This information helps segment users based on the MNO they use, which further helps in the clustering models discussed later. The getSimOperatorName method under the class TelephonyManager is used to collect this information.

\subsubsection{Unique Identity}
Every data set collected from the user needs to be distinguishable from another user’s- this is to avoid multiple data entry from the same device in a short period when it does not provide new information. It also helps track the data associated with the device, while moving. The Internet Protocol (IP) address provides this information, as it is unique to a user over a network at a given point of time. The modules used to collect this are InetAddress and NetworkInterface\cite{ref4}.

\subsection{Cloud Clustering Model}
\subsubsection{K- Means Clustering}
K Means algorithm\cite{ref5} is an iterative vector quantization algorithm that partitions the data set into K predefined distinct non-overlapping subgroups (clusters) where each data point belongs to only one group. It makes the intra-cluster data points as similar as possible while also keeping the clusters as different (far) as possible. Data points are assigned to a cluster such that the sum of the squared distance between the data points and the cluster’s centroid (arithmetic mean of all the data points that belong to that cluster) is at the minimum. The lesser the variation within clusters, the more homogeneous (similar) the data points are within the same cluster. The reasons for the popularity of k-means are ease and simplicity of implementation,
scalability, speed of convergence, and adaptability to sparse data. 

The way K means algorithm works in this project is as follows:
\begin{itemize}

\item Specify number of clusters K (5)
\item Initialize centroids by shuffling data set and randomly selecting K data points for the centroids without replacement, with the dimensions being Latitude, Longitude, and RSSI strength.
\item Iterate until there is no change to the centroids. i.e. assignment of data points to clusters does not change.
\item Compute sum of the squared distance between data points and all centroids.
\item Assign each data point to the closest cluster (centroid) and attribute a normalized network strength integer tag (0-5)
\item Compute the centroids for the clusters by averaging all data points that belong to each cluster.
\end{itemize}
The approach K means follows to solve the problem is called Expectation-Maximization. The E-step is assigning the data points to the closest cluster. The M-step is computing the centroid of each cluster.
The objective function is:
%\begin{figure}[H]
%      \centering
%      \includegraphics[scale=0.4]{images/f1.png}
     
%   \end{figure}
\begin{equation}
J=\sum_{i=1}^{m} \sum_{k=1}^{K} w_{i k}\left\|x^{i}-\mu_{k}\right\|^{2}
\end{equation}
where 
$w_{ik} = 1$ for data point x if it belongs to cluster k; otherwise, $w_{ik} =0$.
$u_{k}$ is the centroid of $x_{i}$’s cluster.

It is a minimization problem of two parts. First J w.r.t. $w_{ik}$ is minimized treating $u_{k}$ as a constant. Then J w.r.t. $u_{k}$ is minimized treating $w_{ik}$ constant. Therefore E step is:
%\begin{figure}[H]
%      \centering
%      \includegraphics[scale=0.32]{images/f2.png}
     
%   \end{figure}
\begin{equation}
\begin{aligned}
\frac{\partial J}{\partial w_{i k}}=\sum_{i=1}^{m} \sum_{k=1}^{K}\left\|x^{i}-\mu_{k}\right\|^{2} & \\
\Rightarrow w_{i k}=\left\{\begin{array}{ll}
1 & \text { if } k=\operatorname{argmin}_{j}\left\|x^{i}-\mu_{j}\right\|^{2} \\
0 & \text { otherwise. }
\end{array}\right.
\end{aligned}
\end{equation}

In other words, assign the data point $x_i$ to the closest cluster judged by its sum of squared distance from cluster’s centroid.
M-step is:
%\begin{figure}[H]
 %     \centering
 %     \includegraphics[scale=0.32]{images/f3.png}
     
%   \end{figure} 
\begin{equation}
\begin{aligned}
\frac{\partial J}{\partial \mu_{k}}=2 \sum_{i=1}^{m} w_{i k}\left(x^{i}-\mu_{k}\right)=& 0 \\
\Rightarrow \mu_{k}=\frac{\sum_{i=1}^{m} w_{i k} x^{i}}{\sum_{i=1}^{m} w_{i k}}
\end{aligned}
\end{equation}
Which translates to recomputing the centroid of each cluster to reflect the new assignments.
%\begin{figure}[H]
%      \centering
%      \includegraphics[scale=0.4]{images/f4.png}
%     
%   \end{figure} 
\begin{equation}
\frac{1}{m_{k}} \sum_{i=1}^{m_{k}}\left\|x^{i}-\mu_{c^{k}}\right\|^{2}
\end{equation}

\subsubsection{Deployment as Cloud Service }
Once the clustering model is functional and accurately models new user data to one of the clusters, it is to be deployed as a cloud service that can run in the back end. It should also update the database with the predicted/clustered mapping of users across areas. The updated database will then be used for the navigation modules discussed later. 
The output of the model will be the cluster predictions denoted by integer tag, along with the geographical locations.
The service is hosted using Google Virtual-Machine\cite{ref6} (VM) Instance. A virtual machine is a software that acts as an interface between a computer program that has been compiled into instructions understood by the virtual machine and the microprocessor (or “hardware platform”) that actually performs the program's instructions.
A Terminal Multiplexer (TMUX\cite{ref7}) session is initialized to allow multiple terminal accessibilities, which is facilitated by the VM instance, using the Secure Shell (SSH\cite{ref8}). TMUX is a protocol
that allows multiple short transport segments, independent of
application type, to be combined between a server and host pair. The Secure Shell (SSH) Protocol is a protocol for secure remote login and other secure network services over an insecure network\cite{ref8}. A Cron-Job\cite{ref9} is used to schedule the service to run as a task every 15 minutes.
Essentially, the clustering on all data- including the latest data that was collected at a 10s interval- is modeled, and the output is updated onto the database. 
This ensures that the changes in the heat map or the general tendencies of variations are always mapped. The 15-minute buffer is optimal- as this is too low a period for massive network property changes to occur, and at the same time gives a sufficient threshold to account for small substantial changes in network details or heat maps, which have to be updated via the clustering model. 

\subsection{Heat map, Nearest Strong Network Area Navigation}
\subsubsection{Heat map and Nearest Strong Network Detection}

The updated database, which contains the geographical location associated with a network strength in the form of normalized tags is presented to the user as a heat map of network strengths. This helps in providing a dynamic visual representation of varying network strength and range across nearby geographical areas, centered around the user. Fig.2 is the heat map of Bangalore generated using sample values of signal strength for illustration purposes.
\begin{figure}[H]
      \centering
      \includegraphics[scale=0.3]{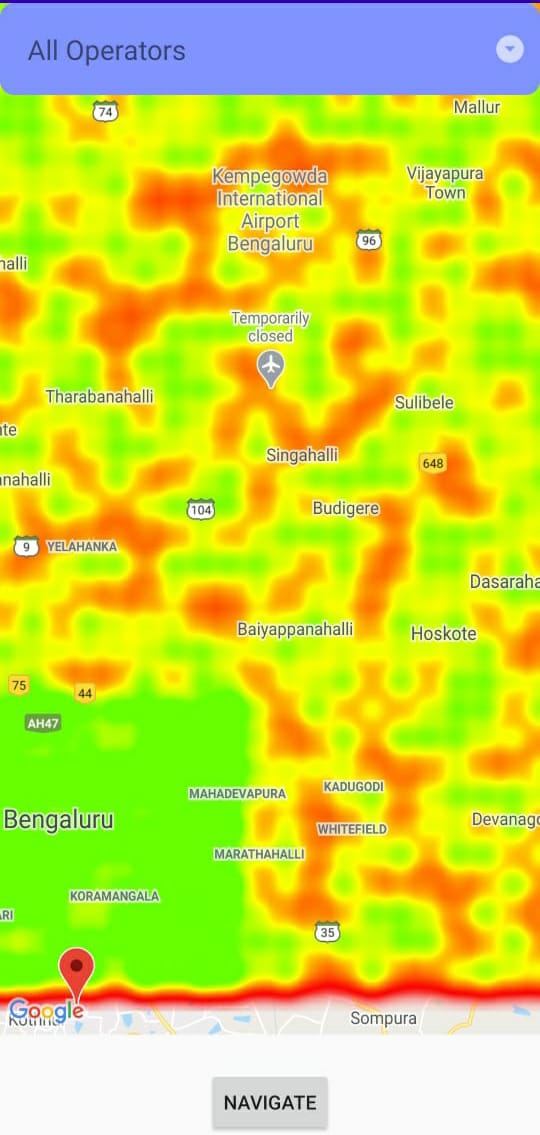}
      \caption{Heat map of Bangalore using sample data for illustration purpose}
      \label{figurelabel3}
   \end{figure}
The database is analyzed with a click of a button “Find Nearest Strong Network”; to find the 3 most optimal nearest stronger network locations to the user, in a 100m radius. The user then can choose the one location which is most comfortable to reach. 

\subsubsection{Navigation}
Given the user’s real-time location, and the location of the nearest optimal stronger network location (corresponding to a higher tag integer, than the one associated with the user’s location), a walkable route can then be traced to reach from one point to the other. This is done using predefined Application Programming Interfaces (API), such as Direction and Routes\cite{ref10} API, provided by Google. In Fig.3, the red marker represents the user's current location and the green marker represents the closest location at which the signal strength is the strongest. The red marker gets updated when there is a change in the user's current location. 
\begin{figure}[H]
      \centering
      \includegraphics[scale=0.3]{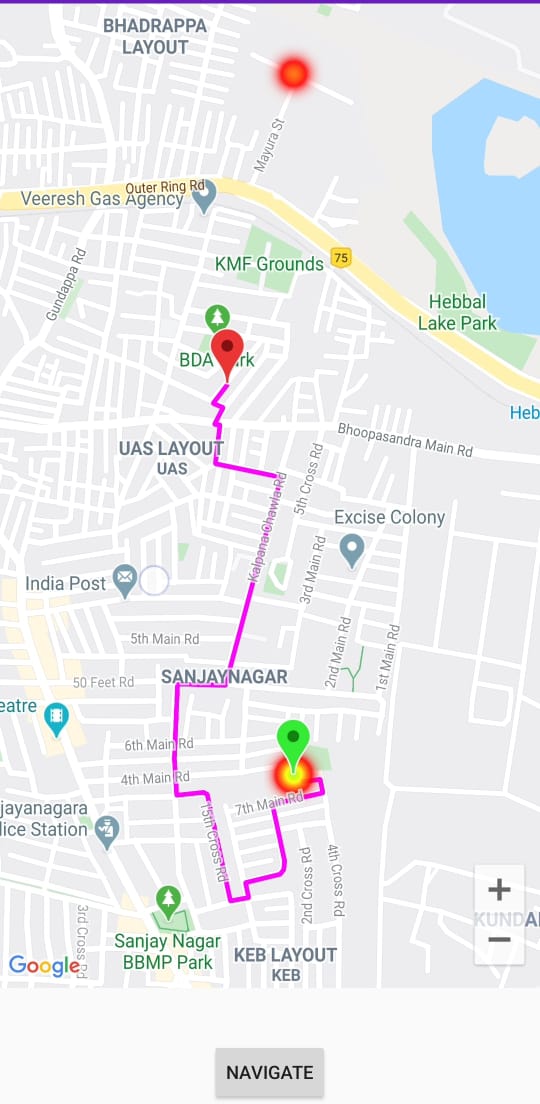}
      \caption{Navigating a user to the nearest green zone}
      \label{figurelabel4}
   \end{figure}

\section{CONCLUSIONS}

A novel approach, to navigate to better mobile network coverage area with stronger signal strength using Machine Learning Clustering Algorithms while deploying it as a Mobile Application has been discussed and tested. This paper helps in understanding the technicalities involved in collecting relevant data using a mobile application, storing them on a real-time database, Firebase, and using a cloud service to cluster the data using the K-means algorithm. The Mobile Application provides a dynamic visual representation of varying network strength and range across nearby geographical areas, centered around the user and also guides the user to a location with the strongest signal strength. 
There is a lot of scope for future work. Analyses beyond heat-map and navigation map can be done with the data collected, to determine the density of the population that resides in low signal strength areas, using DBSCALE\cite{ref12}.
Future work extends to areas of implementing the same idea but by using DENCLUE\cite{ref13} as the clustering model instead of K-Means-Clustering. This could prove to be useful for MNOs looking to increase their connectivity and quality. 

%%%%%%%%%%%%%%%%%%%%%%%%%%%%%%%%%%%%%%%%%%%%%%%%%%%%%%%%%%%%%%%%%%%%%%%%%%%%%%%%

\section*{ACKNOWLEDGMENT}

We would like to thank Mr. Karthik Natarajan for mentoring us. Without his valuable knowledge and expertise, this would not have been possible. We would also like to thank Prof. M Rajasekar (PES University) for providing us this opportunity and for his continuous support. We extend our thanks to PES University, who provided us a platform that helped us to team up and pursue this project.

%%%%%%%%%%%%%%%%%%%%%%%%%%%%%%%%%%%%%%%%%%%%%%%%%%%%%%%%%%%%%%%%%%%%%%%%%%%%%%%%

\bibliographystyle{IEEEtran}
\bibliography{ref}

\end{document}